\documentclass{article} %
\usepackage[final]{neurips_2024}

\usepackage[utf8]{inputenc} %
\usepackage[T1]{fontenc}    %
\usepackage[hidelinks,colorlinks=true,linkcolor=blue,citecolor=blue]{hyperref}       %
\usepackage{hyperref}       %
\usepackage{url}            %
\usepackage{booktabs}       %
\usepackage{amsfonts}       %
\usepackage{nicefrac}       %
\usepackage{microtype}      %
\usepackage{xcolor}         %
\usepackage{enumitem}

\usepackage{amsmath}
\usepackage{amsfonts}
\usepackage{amsthm}
\usepackage{graphicx}

\usepackage{algorithm}
\usepackage{algorithmicx}
\usepackage[noend]{algpseudocode}
\usepackage{std_definitions}
\usepackage{longtable}
\usepackage{bbm}

\newcommand{\protocolfill}[1]{\textcolor{blue}{#1}}

\newcommand{\cost}{\Delta_{\textrm{edit}}}

\newcommand{\framework}{PRELUDE}
\newcommand{\algname}{CIPHER}

\newcommand\cn{$^\clubsuit$}
\newcommand\msr{$^\diamondsuit$}

\usepackage{pgfplots}
\usepackage{filecontents}
\usepackage{subfig}
\usepgfplotslibrary{groupplots}

\usepackage{ulem}   %
\definecolor{oracle}{HTML}{b7b7b7} %
\definecolor{nolearning}{HTML}{000000} %
\definecolor{continual}{HTML}{9900ff} %
\definecolor{ee}{HTML}{9900ff} %
\definecolor{iclbert}{HTML}{f1c232} %
\definecolor{iclmpnet}{HTML}{e06666} %
\definecolor{cotbert}{HTML}{f1c232} %
\definecolor{cotmpnet}{HTML}{e06666} %
\definecolor{bert1}{HTML}{3c78d8}
\definecolor{bert5}{HTML}{3c78d8} %
\definecolor{mpnet1}{HTML}{3CB371} %
\definecolor{mpnet5}{HTML}{3CB371}

\definecolor{darkorange1}{HTML}{e69138} %

\definecolor{context}{HTML}{e69138} %
\definecolor{revision}{HTML}{cc4125} %
\definecolor{response}{HTML}{6aa84f} %
\definecolor{preference}{HTML}{3c78d8} %

\newcommand{\context}[1]{\textbf{\textcolor{context}{#1}}}
\newcommand{\response}[1]{\textbf{\textcolor{response}{#1}}}
\newcommand{\preference}[1]{\textbf{\textcolor{preference}{#1}}}
\newcommand{\revision}[1]{\textbf{\textcolor{revision}{#1}}}

\newcounter{protocol}
\makeatletter
\newenvironment{protocol}[1][htb]{%
  \let\c@algorithm\c@protocol
  \renewcommand{\ALG@name}{Protocol}%
  \begin{algorithm}[#1]%
  }{\end{algorithm}
}
\makeatother

\title{
Aligning LLM Agents by Learning Latent Preference from User Edits
}

\author{
Ge Gao\cn$^*$ \hspace{0.3em} Alexey Taymanov\msr$^*$  \hspace{0.3em} Eduardo Salinas\msr \hspace{0.3em}  Paul Mineiro\msr \hspace{0.3em} Dipendra Misra\msr  \\
\hspace{0.05em} Department of Computer Science, Cornell University\cn \hspace{1.1em} Microsoft Research New York\msr \\
\texttt{ggao@cs.cornell.edu}
\hspace{0.2em} \texttt{\{ataymano, edus, pmineiro, dimisra\}@microsoft.com}  
}

\begin{document}

\maketitle

\renewcommand*\thefootnote{\textbf{$*$}}\footnotetext{Equal contribution.}
\renewcommand*{\thefootnote}{\arabic{footnote}}
\setcounter{footnote}{0}

\begin{abstract}
We study interactive learning of LLM-based language agents based on user edits made to the agent's output. 
In a typical setting such as writing assistants, the user interacts with a language agent to generate a response given a context, and may optionally edit the agent response to personalize it based on their \emph{latent} preference, in addition to improving the correctness. The edit feedback is \emph{naturally generated}, making it a suitable candidate for improving the agent's alignment with the user's preference, and for reducing the cost of user edits over time.
We propose a learning framework, \textbf{\framework}~%
that infers a description of the user's latent preference based on historic edit data. The inferred user preference descriptions are used to define prompts for generating responses in the future. This avoids fine-tuning the agent, which is costly, challenging to scale with the number of users, and may even degrade its performance on other tasks. Furthermore, learning descriptive preference improves interpretability, allowing the user to view and modify the learned preference. However, user preference  can be complex, subtle, and vary based on context, making it challenging to learn. To address this, we propose a simple yet effective algorithm named \textbf{\algname} %
that leverages the LLM to infer the user preference for a given context based on user edits. In the future, \algname~retrieves inferred preferences from the $k$-closest contexts in the history, and forms an aggregate preference for response generation. We introduce two interactive environments -- summarization and email writing, and use a GPT-4 simulated user for evaluation. On both tasks,~\algname~outperforms several baselines 
by achieving the lowest edit distance cost while only having a small overhead in LLM query cost over the base agent. %
Our analysis reports that user preferences learned by~\algname~show significant similarity to the ground truth latent preferences.\footnote{Our code and data are publicly available at \url{https://github.com/gao-g/prelude}.}\looseness=-1

\end{abstract}

\section{Introduction}

Language agents based on large language models (LLMs) have been developed for a variety of applications~\citep{githubcopilot,brynjolfsson2023generative}, following recent breakthroughs in improving LLMs~\citep{achiam2023gpt,ouyang2022training,team2023gemini}. However, despite their impressive zero-shot performance, LLMs still need to align to a given user and task~\citep{mysore2023pearl,li2023automatic}. In many applications, a natural feedback for LLM-based agents is user edits, where a user queries the agent and edits the agent's response before their own final use. In contrast, typical feedback used for fine-tuning, such as the comparison-based preference feedback in RLHF, is explicitly collected by providing annotators with model responses and asking them to rank~\citep[inter alia]{Ziegler2019FineTuningLM,Stiennon2020LearningTS,Nakano2021WebGPTBQ,Ouyang2022TrainingLM}, making such feedback an expensive choice for improving alignment. Motivated by this observation, we focus on interactive learning of LLM-based language agents using user edits as feedback.\looseness=-1

\begin{figure*}[!t]
    \centering
    \vspace{-15pt}
    \includegraphics[clip, trim=1cm 0.5cm 3.5cm 1cm, width=1.00\textwidth]{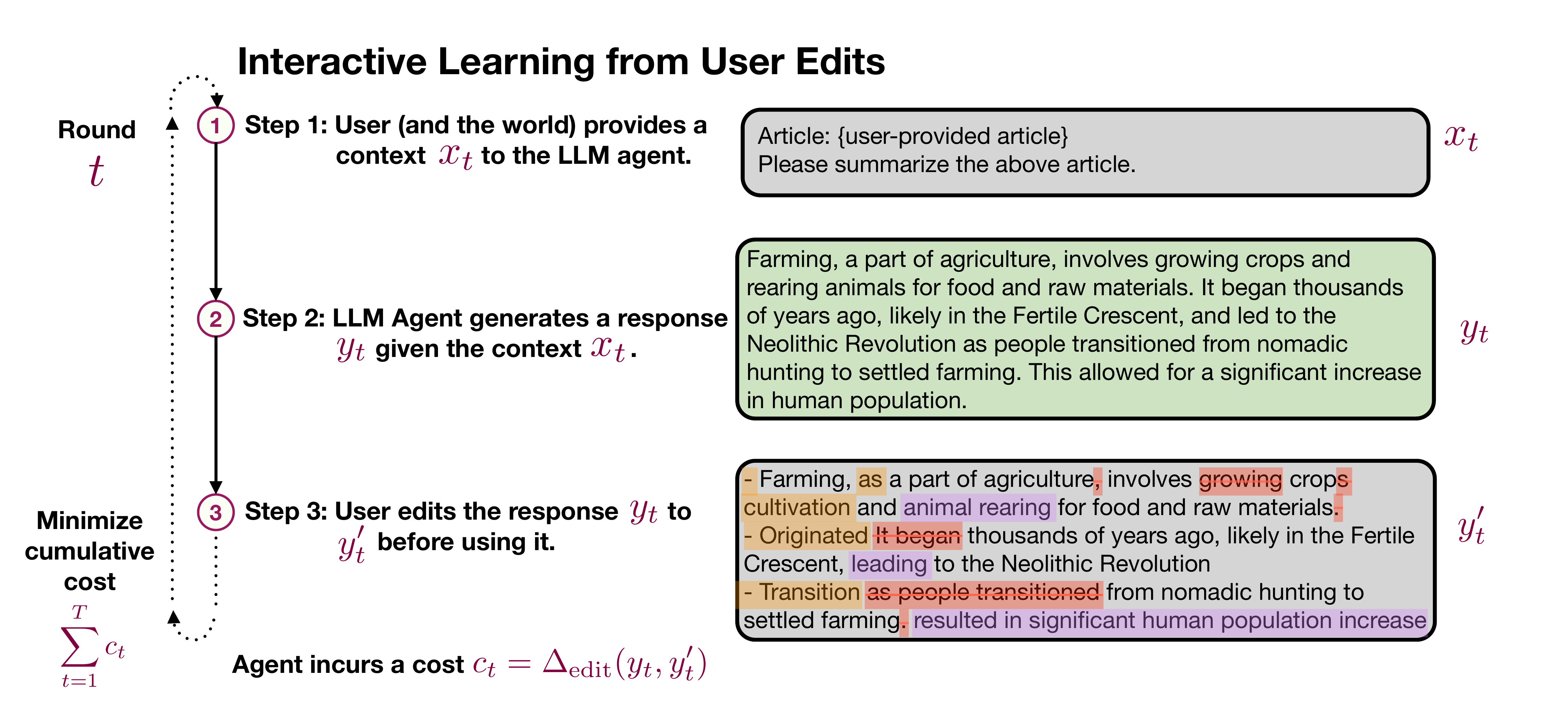} 
    \caption{Illustration of interactive learning from user edits. Color coding in edits is for visualization only -- our agent takes the plain revised text as feedback. }
    \label{fig:main}
    \vspace{-10pt}
\end{figure*}

Consider the scenario in~\pref{fig:main} where a user interacts with an LLM-based writing assistant (agent) to complete their task. The interaction starts with the user (and the world) providing a context to the agent. This context may include a query prompt provided by the user, along with additional information provided by the world, such as the content on the screen, current time, and the user's calendar information. The agent generates a textual response given the context.

In the beginning, the agent's response may not be optimal for the user, as it is not personalized to this user's individual needs and preference. As most users are not familiar with prompt engineering, and LLMs are often able to generate a reasonable response for the task, therefore, users may find it the most convenient to simply edit this response when it is not ideal, rather than trying different prompts to get new responses. The example in~\pref{fig:main} illustrates that the user directly edits the summary generated by the agent to satisfy their preference for bullet point format. It takes time and effort for the user to make edits which can be measured using metrics such as the edit distance between the agent's response and the user edits. Our goal is to minimize the cumulative user edit cost over time using feedback from user edits. Notably, there is no distinction between training and testing in our setting as \emph{every natural use of the agent yields an edit feedback for learning}.\looseness=-1

We conjecture that user edits are driven by user's hidden preference which can be described in natural language. These \emph{preference descriptions} are different from the notion of comparison-based preference used in RLHF. In this paper, we use the word \emph{preference} to mean \emph{preference descriptions}. For instance, preference of the user in~\pref{fig:main} can be described as \emph{bullet points}. In practice, user preference can be compound, such as preferring \emph{bullet point, informal, with emojis} at the same time, and also context-dependent, e.g., \emph{informal} tone when writing an email to a family member, and \emph{formal} tone when writing to a colleague. In more complex settings, user preference can evolve with time (non-stationary), or depend on information unavailable in the context (partially observed). Further, users may not be fully aware of all their preferences, or may fail to express these preferences in their query prompt. 
These considerations imply that user preference is \emph{latent} to the language agent. If the agent could learn the \emph{latent} preference correctly, it can significantly improve its performance by generating satisfactory responses. Furthermore, preference learned by the agent can be shown to the user to enhance \emph{interpretability}, and can even be modified by the user to improve correctness. Motivated by this, we propose a learning framework, \textbf{\framework} (\textbf{PRE}ference \textbf{L}earning from \textbf{U}ser's \textbf{D}irect \textbf{E}dits), where we seek to learn a user preference description for a given context using the history of user edits.\looseness=-1

In a typical real-world scenario such as writing assistants, one has to potentially update the LLM-based agent for every user. Efficient approaches, therefore, must scale with the number of users. This makes approaches that fine-tune LLM parameters expensive to scale. Furthermore, LLMs typically undergo rigorous evaluation on a variety of safety tests before being released, and fine-tuning them can result in loosing the safety guarantees offered by these tests. For example, fine-tuning GPT-4 for millions of users can quickly turn very expensive. Approaches such as adding LORA and Adapter layers and only updating them, or using federated learning, can reduce the expense to some extent, but the loss of safety guarantees remains a concern. In this work, we focus on leveraging a frozen, black-box LLM, and instead learning a \emph{prompt policy} that can infer user preference description for a given context, and then use it to directly drive the response generation.\looseness=-1

We introduce a simple yet effective algorithm~\textbf{\algname}~%
that implements the \framework~framework. 
\algname~infers user preference for every context in the history with the aid of an LLM. In the future, given a context, it retrieves inferred preferences of similar contexts from the history and uses them to generate a response. \algname~is computationally efficient and only  slightly increases the LLM query cost compared to the base agent.\looseness=-1

We introduce two interactive environments that evaluate the agent's ability to summarize documents and compose emails from a given notes. These tasks are inspired by  
writing assistant applications.%
For both tasks, we simulate a GPT-4 user that can generate edits based on a pre-designed \emph{latent} preference that can vary based on the context. 
We evaluate \algname~against several baselines and show that it achieves the lowest user edit cost.
Additionally, \algname~results in a lower LLM query cost than other retrieval-based baselines.
Finally, we %
analyze preferences learned by our agents, and find that they show significant similarity to the ground truth latent preferences in our setup.\looseness=-1

\vspace*{-0.2cm}
\section{Interactive Learning from User Edits and the \framework~Framework}
\vspace*{-0.1cm}
We first describe LLM agents and the general learning framework from user edits and then discuss our \framework~framework and associated learning challenges.%

 \noindent\textbf{LLM and Language Agents.} We assume access to a language agent that internally relies on an LLM. We make no assumption on the agent except that it can take as input a piece of context which can include both texts and images and an additional prompt (which can be in-context learning examples or learned preferences) and generates a text response. The language agent may simply perform greedy decoding of the LLM given the input or may perform complex planning to generate a response.

\vspace{-5pt}
\begin{protocol}
\caption{\textbf{Interactive Learning from User Edits}.}
\label{proto:learning_from_edits}
    \begin{algorithmic}[1]
        \For{$t=1, 2, \cdots, T$}
        \State User and the world provide a context $x_t$
        \State \protocolfill{Agent generates a response $y_t$ given the context $x_t$}
        \State User edits the response to $y'_t$
        \State Agent receives a cost of $c_t = \cost(y_t, y'_t)$
        \EndFor
        \State Evaluate the agent and learning algorithm on $\sum_{t=1}^T c_t$
    \end{algorithmic}
    \vspace*{-0.1cm}
\end{protocol}
\vspace{-5pt}

 \noindent\textbf{Interactive Learning from User Edits.} In an application such as a writing assistant, a user interacts with the language agent over $T$ rounds. \pref{proto:learning_from_edits} shows such learning protocol. In the $t^{th}$ round, the user and the world provide a context $x_t \in \Xcal$ where $\Xcal$ is the space of all possible contexts. This context will include the user prompt in text, along with additional information provided by the user or the world, and may include multimodal data as well such as images. Given the context $x_t$, the language agent generates a response $y_t \in \Ycal$ in text, where $\Ycal$ is the space of all texts. The user edits the response $y_t$ to $y'_t$. If the user does not perform any edits, we treat this as setting $y'_t=y_t$. The agent receives a cost of $c_t = \cost(y_t, y'_t)$ for this round, which measures the user's efforts on making edits. The goal of the agent is to minimize the sum of costs across all rounds $\sum_{t=1}^T c_t$. In our experiments, we use $\cost$ as Levenshtein edit distance~\citep{Levenshtein1965BinaryCC} in the token space which computes the minimum number of token insertion, deletion, and substitution necessary to convert $y_t$ to $y'_t$. In general, a higher edit distance implies that the user has made more edits and spent more efforts. %

 \noindent\textbf{\framework~Framework.} We describe our \framework~framework in~\pref{proto:framework} which is a specialization of%
 ~\pref{proto:learning_from_edits}. In \framework, in the $t^{th}$ round, the agent infers the preference of the user as $f_t$, and uses it to generate a response. We assume that in this round and for the given context $x_t$, the user has a \emph{latent} preference $f^\star_t$ that drives the user to perform all edits. Furthermore, we assume that if the agent was able to infer this \emph{latent} preference ($f_t = f^\star_t$), then it will lead to minimal possible edits.\footnote{The edit cost in practice may not always be 0, as the language agent could be incapable of adeptly using the correct preference, or the user may perform edits that are inconsistent with their preference.} To remove the dependence on performance due to the choice of the base LLM agent, we compare with an oracle agent that has access to $f^\star_t$ at the start of each round. We assume that the LLM remains frozen across all methods in this work.

\begin{protocol}[t!]
\vspace*{-0.1cm}
\caption{\textbf{\framework}: \textbf{PRE}ference \textbf{L}earning from \textbf{U}ser's \textbf{D}irect \textbf{E}dits}
    \begin{algorithmic}[1]
        \For{$t=1, 2, \cdots, T$}
        \State User presents a text context $x_t$
        \State \protocolfill{Agent infers a preference $f_t$ using the history $\left\{(x_\ell, y_\ell, y'_\ell)\right\}_{\ell=1}^{t-1}$ and context $x_t$} 
        \State \protocolfill{Agent uses $f_t$ and $x_t$ to generate a response $y_t$}
        \State User edits the response to $y'_t$ using their \emph{latent} preference $f^\star_t$
        \State Agent incurs a cost $c_t = \Delta(y_t, y'_t)$
        \EndFor
        \State Return $\sum_{t=1}^T c_t$
    \end{algorithmic}
    \label{proto:framework}
    \vspace*{-0.1cm}
\end{protocol}

 \noindent\textbf{Challenges of Learning User Preference.} Learning user preference from edits is challenging. In practice, user preference are multifaceted and complex. Furthermore, user's preference can also significantly vary based on the context. The feedback in the form of user edits emerges naturally but is inherently implicit, lacking direct expressions of the actual preference and carrying subtleties that may lead to diverse interpretations. The combination of preference variability and the implicit nature of feedback poses considerable challenges for agents in accurately learning and integrating these preferences.\looseness=-1

\section{Learning User Preference through Retrieval and Aggregation}

In this section, we present our method,~\algname~(\textbf{C}onsolidates \textbf{I}nduced \textbf{P}references based on \textbf{H}istorical \textbf{E}dits with \textbf{R}etrieval), that learns user preference based on user edits.

\vspace{-5pt}
\begin{algorithm}[h!]
\caption{\algname$(\phi, k, \delta)$. A context representation function $\phi: \Xcal \rightarrow \RR^d$, the retrieval hyperparameter $k$, and tolerance hyperparameter $\delta \ge 0$. We initialize history $\Dcal = \emptyset$.}
    \begin{algorithmic}[1]
        \For{$t=1, 2, \cdots, T$}
        \State User (and the world) presents a context $x_t$
        \State Retrieve the top-$k$ examples $\{\phi(x_{z_i}), \tilde{f}_{z_i}\}_{i=1}^k$ in $\Dcal$ with maximum cosine similarity to $\phi(x_t)$
        \State If $k > 1$, then query the LLM to aggregate these preferences $\{\tilde{f}_{z_i}\}_{i=1}^k$ into $f_t$, else $f_t = \tilde{f}_{z_1}$\label{line:merge_preferences}
        \State Agent generates a text response $y_t$ based on  $x_t$ and $f_t$ \label{line:generate}
        \State User edits the response to $y'_t$ using their latent preference $f^\star_t$
        \State Agent incurs a cost $c_t = \cost(y_t, y'_t)$
        \If{$c_t \le \delta$}
        \State $\tilde{f}_t = f_t$
        \Else
        \State Query the LLM to generate a preference $\tilde{f}_t$ that best explains user edits in $(y_t, y'_t)$\label{line:infer_preference}
        \EndIf
        \State $\Dcal \leftarrow \Dcal \cup \{(\phi(x_t), \tilde{f}_t)\}$
        \EndFor
        \State Return $\sum_{t=1}^T c_t$
    \end{algorithmic}
    \label{alg:cipher}
\end{algorithm}
\vspace{-5pt}

\pref{alg:cipher} shows \algname~which implements the \framework~framework. \algname~maintains a preference history $\Dcal_t = \{(x_\ell, \tilde{f}_\ell)\}_{\ell=1}^{t-1}$ of past contexts $x_\ell$ along with a preference $\tilde{f}_\ell$ inferred by the agent. \algname~assumes access to a \emph{context representation function} $\phi:\Xcal \rightarrow \RR^d$ that can map a context to a vector representation. For a given round $t$ with context $x_t$, the agent first retrieves the $k$-closest contexts from the interaction history $\Dcal_t$. We use cosine similarity for computing proximity, although other metrics such as Euclidean distance, or Hamming distance when $\phi$ outputs a binary vector, can be used. Given the retrieved contexts and their inferred preferences $\{(x_{z_i}, \tilde{f}_{z_i})\}_{i=1}^k$, we query the underlying LLM to summarize the inferred preferences $\{\tilde{f}_{z_i}\}_{i=1}^k$ into a single preference $f_t$. In the beginning, when $t \le k$, we retrieve all the past $t$ contexts. In particular, for $t=1$ we have $f_1$ as an empty string as the agent has no prior knowledge of this user's preference.\footnote{In practice, one can initialize with a publicly available preference history.}

The agent uses the inferred preference $f_t$ to generate the response. This is done by concatenating the context $x_t$ with an agent prompt such as ``\emph{This user has a preference of <$f_t$> which must be used when generating the response}'', where <$f_t$> indicates where we insert the inferred preference $f_t$. We list the actual template used in our experiments in~\pref{tab:agent_prompt_template} in~\pref{app:addition_details}.

Given the user edits $y'_t$, if the user edits are minimal, i.e., $\cost(y_t, y'_t) \le \delta$ for a hyperparameter $\delta$, then we 
set the inferred preference for this round as $\tilde{f}_t = f_t$
 as using $f_t$ for generating a response resulted in minimal edits. However, if $\cost(y_t, y'_t) > \delta$, then we query the LLM a third time to generate the inferred preference $\tilde{f}_t$ that explains why the user edited $y_t$ to $y'_t$.  We call this the \emph{Latent Preference Induction} (LPI) step. In both cases, we append $(x_t, f_t)$ to the preference history.
 
 Note that we cannot query the LLM for the inferred preference in the first case where the user edit cost $c_t$ is small, i.e., $c_t \le \delta$. In this case, querying the LLM to infer the preference to explain the edits in $y'_t$ given $y_t$, will result in the LLM outputting that the agent has no preference which is incorrect. %

 \noindent\textbf{Computational Cost of \algname.} In a given round, \algname~adds a maximum of 3 LLM calls on top of the cost of calling the underlying inference algorithm of the agent in~\pref{line:generate}. \algname~further reduces the memory storage by only storing the representation of contexts in the preference string instead of the input itself. Finally, \algname~only adds a small prompt to the context $x_t$, before calling the agent's inference algorithm. This only slightly increases the length of the prompt, thereby, reducing the query cost associated with LLMs that scales with the number of input tokens.

\section{Experiment}
We first introduce two interactive tasks for learning from user edits, and then describe our results.

\subsection{Two Interactive Writing Assistant Environments for Learning from User Edits}
 \noindent\textbf{Task.} We introduce two tasks inspired by the use of LLMs as writing assistants~\citep{mysore2023pearl,shen2023beyond,wang2023writing}. In the first task, we evaluate the agent's ability to summarize a given document. In the second task, we evaluate the agent's ability to compose an email given notes. For both tasks, we use documents from several existing sources listed in~\pref{tab:latent_user_pref}. These sources represent a diverse category of documents that a writing assistant would typically encounter (see~\autoref{tab:context_link_example} in Appendix for examples).  %
In any given round, the user is provided a context that is a document from one of the sources for the given task. Importantly, the agent is \emph{unaware of the source of the given document} which as we discuss later, will determine the user preference. For both tasks, we run an experiment for $T = 200$ rounds. We sample an equal number of documents from each source and mix them to remove any temporal correlation in document sources.

\begin{table}[t!]
    \centering \small
    \setlength{\tabcolsep}{4.5pt}
    \caption{Latent user preference design, specific to the document source.}   
    \begin{tabular}{p{0.2\linewidth} p{0.43\linewidth} p{0.3\linewidth}}
        \toprule
        \textbf{Doc Source} & \textbf{Latent User Preference} & \textbf{Scenario} \\
        \midrule
        \textbf{Summarization} & &  \\
         News article\newline\citep{see-etal-2017-get} & targeted to young children, storytelling, short sentences, playful language, interactive, positive & introduce a political news to kids \\
         Reddit post\newline\citep{Stiennon2020LearningTS} & second person narrative, brief, show emotions, invoke personal reflection, immersive & for character development in creative writing \\
         Wikipedia page\newline\citep{wikidump} & bullet points, parallel structure, brief & take notes for key knowledge \\
         Paper abstract\newline\citep{clement2019arxiv} & tweet style, simple English, inquisitive, skillful foreshadowing, with emojis & promote a paper to invoke more attention and interests \\
         Movie review\newline\citep{maas-EtAl:2011:ACL-HLT2011} & question answering style, direct, concise & quickly get main opinions \\
        \midrule
         \textbf{Email Writing} & &  \\
         Personal problem\newline\citep{Stiennon2020LearningTS} &  informal, conversational, short, no closing & share life with friends  \\
         Paper review\newline\citep{hua-etal-2019-argument} & casual tone, positive, clear, call to action & peer review to colleague \\
         Paper tweet\newline\citep{Bar_PaperTweet} & engaging, personalized, professional tone, thankful closing & networking emails for researchers \\
         Paper summary\newline\citep{Kershaw2020ElsevierOC} & structured, straight to the points, respectful, professional greeting and closing & milestone report to superiors \\
        \bottomrule
    \end{tabular} 
    \label{tab:latent_user_pref}
    \vspace*{-0.5cm}
\end{table}

 \noindent\textbf{Two-Stage GPT-4 Simulated User.} We simulate a user that can edit a given response. We define a set of \emph{latent user preferences} for the user that vary based on the document source.~\pref{tab:latent_user_pref} lists the preference for every source. This captures the context-dependent nature of user preferences as the document source influences the type of context. For example, the \emph{Personal problem} document source contains documents pertaining to discussions with a friend, and a user may have a different preference when writing an email to a friend compared to writing an email to a colleague. %
 We assume that our user is aware of the document source $d_t$ of a given context $x_t$. This implies, that we can express the true user preference for $x_t$ as $f^\star_t = F(d_t)$ where $F$ maps a given document source to the user preference. Recall that the \emph{agent is never provided the document source of any context}.

We model our user using GPT-4 with a two-stage approach. Given an agent response $y_t$ and the context $x_t$, we first query GPT-4 to check if $y_t$ satisfies the preference in $f^\star_t$. If the answer is yes, then the user preforms no edits and returns $y'_t = y_t$. If the answer is no, then we use GPT-4 to generate the edited response $y'_t$ given $y_t$ and $f^\star_t$. We found that our two-stage GPT-4 user can generate high-quality edits, consistent with observations in prior work that LLM-written feedback is high-quality and useful to learn from~\citep{Bai2022ConstitutionalAH,Saunders2022SelfcritiquingMF}. We adopted a two-stage process since using GPT-4 to directly edit the response $y_t$ always resulted in edits even when the response satisfied the preference $f^\star_t$. We provide GPT-4 user prompt template and user edit examples in~\pref{app:addition_details}.\looseness=-1

 \noindent\textbf{Evaluation Metric.} We propose three metrics for evaluating agents learning from user edits. Our main metric is the cumulative user edit cost $\sum_{t=1}^T \cost(y_t, y'_t)$ over $T$ rounds where $\cost(y_t, y'_t)$ is the Levenshtein edit distance between agent response $y_t$ and user edits $y'_t$ computed in the token space using Tiktoken tokenizer.
For methods that learn an interpretable preference,
we additionally evaluate the quality of the inferred user preference $f_t$. We do so by evaluating if $f_t$ is closer to the true preference $f^\star_t = F(d_t)$, where $d_t$ is the document source of context in round $t$, compared to preference of any other document source. Formally, we compute~$\frac{1}{T}\sum_{t=1}^{T} \mathbbm{1}\{d_t = \arg\max_{d \in \Scal} \text{BERTScore}(f_t, F(d))\}$, where \text{BERTScore}~\citep{bert-score} is a text similarity metric and $\Scal$ is the set of all document sources.
Finally, we report the total number of input and output BPE tokens to the LLM across all rounds. This measures the expense associated with using LLM, used by popular LLM providers to charge their customers.

\subsection{Details of \algname~and Comparison Systems}

We use GPT-4 as our base LLM for \algname~and all baselines. We do not perform fine-tuning of the GPT-4 and do not add any additional parameters to the model. We use a prompt-based GPT-4 agent for all methods that uses a single prompt with greedy decoding to generate the response. Our main method \algname~and the baselines, can be extended to more complex language agents that perform multiple steps of reasoning on top of the base LLM before generating a response.

 \noindent\textbf{\algname~Details.} We use a simple agent that uses GPT-4 with a prompt template to generate the response $y_t$ given the context $x_t$ and preference $f_t$. We list templates in~\pref{tab:agent_prompt_template} in~\pref{app:addition_details}. We experiment with MPNET~\citep{Song2020MPNetMA} and BERT~\citep{Devlin2019BERTPO} as our two context representation functions $\phi$, and use cosine similarity for retrieval. We experiment with two different values of the number of retrieved examples $k \in \{1, 5\}$.

 \noindent\textbf{Baselines.} We evaluate \algname~against baselines that either perform no learning, or learn context-agnostic preferences, or directly use past edits to generate a response:

\vspace*{-0.1cm}
\begin{enumerate}[leftmargin=0.5cm]
    \item \textit{No learning:} The agent performs no learning based on interaction with the user. %

    \item \textit{Explore-then-exploit (E-then-e) LPI:} This baseline is based on the classic explore-then-exploit strategy in interactive learning~\citep{garivier2016explore}. The agent first generates responses for the first $T_e$ rounds without performing any learning (exploration stage). It then infers a single user preference $\tilde{f}_e$ using the user edits in the first $T_e$ rounds by applying the LPI step (\pref{alg:cipher}, \pref{line:infer_preference}), which is used to generate responses for remaining rounds (exploitation step).\looseness=-1

    \item \textit{Continual LPI:} This baseline is similar to \textit{E-then-e LPI} except that it never stops exploring and avoids overfitting to the first $T_e$ rounds. In any given round $t$, it uses the data of all past edits~$\{(y_\ell, y'_\ell)\}_{\ell=1}^{t-1}$ to learn a preference $f_t$ by performing the LPI step. It then generates a response using this preference. Similar to \textit{E-then-e LPI}, this approach learn context-agnostic preferences.\looseness=-1
    \item \textit{ICL-edit:} This is a standard retrieval-based in-context learning (ICL) baseline~\citep{brown2020language}. In a given round $t$, the agent first retrieves the closest $k$ examples $\{(y_{z_\ell}, y'_{z_\ell})\}_{\ell=1}^k$ to the given context $x_t$ using the representation function $\phi$. These examples are provided in an ICL prompt and use to generate the response $y_t$. This approach does not learn preferences but unlike E-then-e LPI and Continual LPI it can perform context-dependent learning.
     \item \textit{CoT-edit:} This is a standard retrieval-based chain-of-thought (CoT) baseline~\citep{Wei2022ChainOT}. This baseline is similar to \textit{ICL-edit} except the prompt for generation requires the agent to infer a user preference $f_t$ based on retrieved $k$ examples, and generate an output according to $f_t$.\footnote{\autoref{app:addition_details} reports additional details of our baselines, such as hyperparameters and prompt templates.}
\end{enumerate}

\noindent\textbf{Oracle Method.} We also evaluate an oracle approach which uses the true user preference in each round to generate the response. This provides an upper bound on performance and helps to evaluate if our setup is well-designed, i.e., whether learning the true user preference indeed leads to low edit costs.\looseness=-1

\subsection{Main Result and Discussion.}

\noindent\textbf{Main Results.}
\autoref{tab:pfm} reports the performance of all methods on the two tasks on three metrics.
We report the mean and standard deviation across 3 different random seeds.%
\footnote{We randomize the context sampling from source datasets, so experiments on different seeds contain different sets of input contexts. On the same seed, experiments across different methods are strictly comparable, as both the set of input contexts and the order of input context seen are the same in our implementation.} %

\begin{table}[h!]
\vspace{-13pt}
\centering \small
\setlength{\tabcolsep}{5pt}
\caption{Performance of baselines and our methods in terms of cumulative edit distance cost and classification accuracy. $\mu_\sigma$ denotes the mean $\mu$ and standard deviation $\sigma$ across 3 runs over different seeds. Expense column shows budget as the average number of input and output BPE tokens across 3 runs (unit is $\cdot 10^5$). We use \textit{-k} in method names to denote that we use $k$ retrieved examples. Numbers in bold are the best performance in each column excluding \textit{oracle preference} method, underline for the second best, and dotted underline for the third best.}
\begin{tabular}{l c c c c c c }
\toprule
    \textbf{Method} & \multicolumn{3}{c}{\textbf{Summarization}} & \multicolumn{3}{c}{\textbf{Email Writing}} \\
    & Edit Distance$\downarrow$ & Accuracy$\uparrow$ & Expense$\downarrow$ & Edit Distance$\downarrow$ & Accuracy$\uparrow$ & Expense$\downarrow$ \\
\midrule
    Oracle Preference & \hspace{2pt} 6,573\textsubscript{1,451} & 1.000 & 1.67 &  1,851\textsubscript{243}  & 1.000 & 1.62 \\
\midrule
    No Learning &  48,269\textsubscript{957} \hspace{2pt} & - & 1.50 & 31,103\textsubscript{900} \hspace{2pt}  & - & 1.65 \\
    E-then-e LPI & \hspace{2pt} 65,218\textsubscript{17,466} &  0.218\textsubscript{0.003} & 1.99 & 24,562\textsubscript{1,022} & 0.263\textsubscript{0.003} & 1.73 \\
    Continual LPI & 57,915\textsubscript{2,210} & 0.233\textsubscript{0.010} & 8.89 & 26,852\textsubscript{1,464} & 0.243\textsubscript{0.019} & 8.63 \\ %
\midrule
    ICL-edit-5-MPNET & 38,560\textsubscript{1,044} &  - & 8.00 & 32,405\textsubscript{1,307} & - & 12.12 \\
    ICL-edit-5-BERT & 39,734\textsubscript{1,929} & - & 7.96 & 30,949\textsubscript{3,250} & - & 11.55 \\
    CoT-edit-5-MPNET & 40,747\textsubscript{1,874} & 0.230\textsubscript{0.026} & 6.82 &  24,292\textsubscript{3,503} & 0.300\textsubscript{0.023} & 8.74 \\
    CoT-edit-5-BERT & 41,088\textsubscript{1,846} & 0.230\textsubscript{0.013} & 6.92 & 24,301\textsubscript{1,382} & 0.263\textsubscript{0.032} & 8.26 \\
\midrule
    CIPHER-1-MPNET & \underline{33,926}\textsubscript{4,000} & \underline{0.520}\textsubscript{0.022} & 2.74 & \dotuline{10,781}\textsubscript{1,711} & \dotuline{0.435}\textsubscript{0.084} & 1.94 \\
    CIPHER-5-MPNET & \textbf{32,974}\textsubscript{195} \hspace{2pt} & \dotuline{0.478}\textsubscript{0.010} & 3.00 & \underline{10,058}\textsubscript{1,709} & \underline{0.467}\textsubscript{0.081}& 2.09 \\
    CIPHER-1-BERT & 37,637\textsubscript{3,025} & \textbf{0.565}\textsubscript{0.053} & 2.81 & 12,634\textsubscript{4,868}& \textbf{0.487}\textsubscript{0.125}& 1.99 \\
    CIPHER-5-BERT & \dotuline{35,811}\textsubscript{3,384} & \dotuline{0.478}\textsubscript{0.028} & 3.03 & \hspace{2pt} \textbf{8,391}\textsubscript{3,038} & 0.363\textsubscript{0.075}& 2.22 \\
\bottomrule
\end{tabular}
    \label{tab:pfm}
    \vspace{-8pt}
\end{table}

 \noindent\textbf{Discussion of Main Result.} %
We observe that not performing learning results in a high edit cost, whereas using the oracle preferences achieves a significantly smaller edit cost. This shows that our environments are sound and well-conditioned. \textit{E-then-e LPI} and \textit{Continual LPI} learn context-agnostic preferences which cannot capture the context-dependent preferences in the environments and end up doing poorly. For the summarization task, they end up with a higher edit distance than even performing no learning. One possible explanation is that using context-agnostic preferences can push the model to specialize to a given preference much more than the base model, resulting in more edits when that preference is incorrect. We see this in preference accuracy, which is low for both of these baselines, and lower for the summarization task than the email writing task where they outperform no learning baselines. Further, \textit{Continual LPI} has a higher expense cost due to constantly querying the LLM to infer the user preference.\looseness=-1

\textit{ICL-edit} baselines perform significantly better on the summarization task. However, using a list of user edits in the prompt results in a higher token expense cost, as the responses and their edits can be significantly long in practice. Further, the ICL-edit baselines provide no interpretable explanation for their response or for explaining user behavior. Although \textit{CoT-edit} baselines provide an interpretable preference, they still result in relatively high expense and low classification accuracy.\looseness=-1

\algname~achieves the smallest edit distance cost reducing edits by 31\% in the summarization task and 73\% in the email writing task. We observe that retrieving $k=5$ preferences and aggregating them achieves lower edit distance, however, the choice of ideal representation $\phi$ seems task-dependent. Further, \algname~achieves the highest preference accuracy showing that \algname~can learn preferences that correlate more with the ground truth preference than preferences of other document sources. Note that the performance of a random preference classifier is only 20\% for summarization and 25\% for email writing. Further, \algname~achieves a smaller cost than \textit{ICL-edit} and \textit{Continual LPI} baselines, as it doesn't use long user edits in the prompt for generating a response. In summary, \algname~provides a cheap, more effective, and interpretable method than our baselines.\looseness=-1

\begin{figure*}[t!]
\centering \small
\caption{Learning curves of different methods based on cumulative cost over time (average across 3 seeds). In the legend, \textit{-k} means with top $k$ retrieved examples, \textit{-B} for BERT, and \textit{-M} for MPNET.}
\vspace{-8pt}
\begin{tikzpicture}
\begin{groupplot}[
       group style={
          group size=2 by 1,
          horizontal sep=100pt,
          },
       width=.45\textwidth,
       height=.45\textwidth,
       ymin=10, 
       xmin=0, xmax=200, xtick distance=40],
       \nextgroupplot[
           title=\textbf{Summarization},title style={yshift=-1.5ex},
           ylabel={Cumulative Cost}, ylabel style={yshift=-5ex},
           ymin=50,
           xlabel={Round},xlabel style={yshift=1ex},
           legend style={at={(1.55, 0.53)},
           anchor=east, 
           legend cell align={left},
           font=\scriptsize}
       ]
            \addplot [oracle, thick, smooth] table [x=idx, y=oracle, col sep=space] {summarization_cum_cost.csv};
            \addplot [nolearning, densely dotted, thick, smooth] table [x=idx, y=no, col sep=space] {summarization_cum_cost.csv};
            \addplot [ee, loosely dotted, thick,  smooth] table [x=idx, y=ee, col sep=space] {summarization_cum_cost.csv};
            \addplot [continual, dotted, thick, smooth] table [x=idx, y=continual, col sep=space] {summarization_cum_cost.csv};
            \addplot [iclbert, thick, smooth] table [x=idx, y=icl5bert, col sep=space] {summarization_cum_cost.csv};
            \addplot [iclmpnet, thick, smooth] table [x=idx, y=icl5mpnet, col sep=space] {summarization_cum_cost.csv};
            \addplot [cotbert, densely dashed, smooth] table [x=idx, y=Scot5bert, col sep=space] {cot_cum_cost.csv};
            \addplot [cotmpnet, densely dashed, smooth] table [x=idx, y=Scot5mpnet, col sep=space] {cot_cum_cost.csv};
            \addplot [bert1, dashdotted, thick, smooth] table [x=idx, y=bert1, col sep=space] {summarization_cum_cost.csv};
            \addplot [bert5, thick, smooth] table [x=idx, y=bert5, col sep=space] {summarization_cum_cost.csv};
            \addplot [mpnet1, dashdotted, thick, smooth] table [x=idx, y=mpnet1, col sep=space] {summarization_cum_cost.csv};
            \addplot [mpnet5, thick, smooth] table [x=idx, y=mpnet5, col sep=space] {summarization_cum_cost.csv};

            \addlegendimage{oracle}
            \addlegendentry{Oracle}
            \addlegendimage{nolearning, densely dotted}
            \addlegendentry{No Learning}
            \addlegendimage{ee, loosely dotted}
            \addlegendentry{E-then-e}
            \addlegendimage{continual, dotted}
            \addlegendentry{Continual}
            \addlegendimage{iclbert}
            \addlegendentry{ICL-edit-B}
            \addlegendimage{iclmpnet}
            \addlegendentry{ICL-edit-M}
            \addlegendimage{cotbert}
            \addlegendentry{CoT-edit-B}
            \addlegendimage{cotmpnet}
            \addlegendentry{Cot-edit-M}
            \addlegendimage{bert1, dashed}
            \addlegendentry{CIPHER-1-B}
            \addlegendimage{bert5}
            \addlegendentry{CIPHER-5-B}
            \addlegendimage{mpnet1, dashdotted}
            \addlegendentry{CIPHER-1-M}
            \addlegendimage{mpnet5}
            \addlegendentry{CIPHER-5-M}

       \nextgroupplot[
           title=\textbf{Email Writing},title style={yshift=-1.5ex},
           ylabel={Cumulative Cost}, ylabel style={yshift=-5ex},
           xlabel={Round},xlabel style={yshift=1ex},
           legend cell align={left},
           legend style={at={(0.45, 0.71)}, anchor=east, 
           font=\scriptsize}
       ]
            \addplot [oracle, thick, smooth] table [x=idx, y=oracle, col sep=space] {email_cum_cost.csv};
            \addplot [nolearning, densely dotted, thick, smooth] table [x=idx, y=no, col sep=space] {email_cum_cost.csv};
            \addplot [ee, loosely dotted, thick,  smooth] table [x=idx, y=ee, col sep=space] {email_cum_cost.csv};
            \addplot [continual, dotted, thick, smooth] table [x=idx, y=continual, col sep=space] {email_cum_cost.csv};
            \addplot [iclbert, thick, smooth] table [x=idx, y=icl5bert, col sep=space] {email_cum_cost.csv};
            \addplot [iclmpnet, thick, smooth] table [x=idx, y=icl5mpnet, col sep=space] {email_cum_cost.csv};
            \addplot [cotbert, densely dashed, smooth] table [x=idx, y=Ecot5bert, col sep=space] {cot_cum_cost.csv};
            \addplot [cotmpnet, densely dashed, smooth] table [x=idx, y=Ecot5mpnet, col sep=space] {cot_cum_cost.csv};
            \addplot [bert1, dashdotted, thick, smooth] table [x=idx, y=bert1, col sep=space] {email_cum_cost.csv};
            \addplot [bert5, thick, smooth] table [x=idx, y=bert5, col sep=space] {email_cum_cost.csv};
            \addplot [mpnet1, dashdotted, thick, smooth] table [x=idx, y=mpnet1, col sep=space] {email_cum_cost.csv};
            \addplot [mpnet5, thick, smooth] table [x=idx, y=mpnet5, col sep=space] {email_cum_cost.csv};

\end{groupplot}
\end{tikzpicture}
\label{fig:cumulative_cost}
\vspace{-25pt}
\end{figure*}
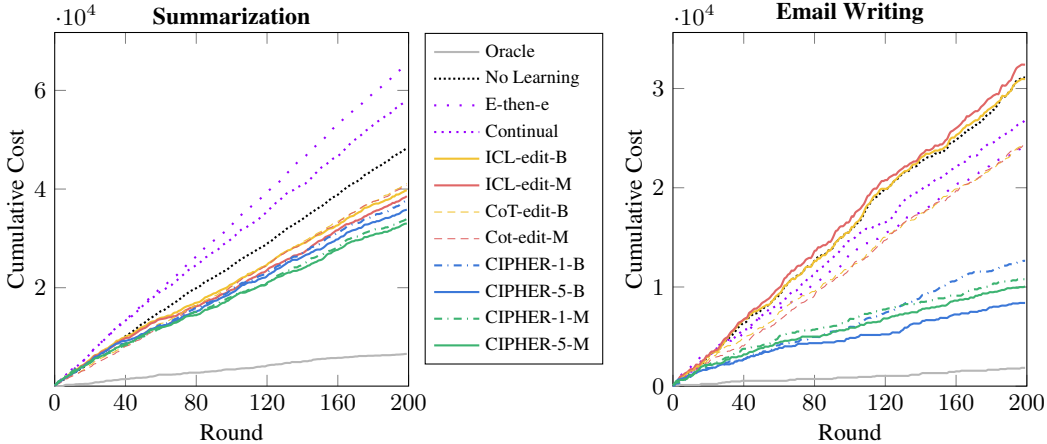

\begin{figure*}[t!]
\centering \small
\caption{%
Percentage of zero-cost examples of CIPHER over time, binned per 20 rounds to show the trend (average across 3 seeds). In the legend, \textit{-k} means with top $k$ retrieved examples, \textit{-B} for BERT, and \textit{-M} for MPNET.\looseness=-1}
\vspace{-10pt}
\begin{tikzpicture}
\begin{groupplot}[
       group style={
          group size=2 by 2,
          vertical sep=25pt,
          horizontal sep=110pt,
          x descriptions at=edge bottom},
       ylabel style={align=center, text width=.22\textwidth},
       width=.4\textwidth,
       height=.27\textwidth],

        \nextgroupplot[
              title=\textbf{Summarization},title style={yshift=-1.5ex},
              ylabel={\% Zero-cost Ex. per Bin}, ymin=0, ymax=0.9,
              ylabel style={yshift=-2.5ex},
              xlabel={Round},xmin=0, xmax=9, xtick distance=1.8, xticklabels={-40,,40,80,120,160,200},
              xlabel style={yshift=1ex},
              yticklabels={,0,0.2,0.4,0.6,0.8,1},
              legend cell align={left},
              legend style={at={(1.05, 0.65)}, 
              anchor=west, font=\scriptsize,, legend columns=1},
              ]
            \addplot [bert5, mark=o, thick, smooth] table [x=idx, y=bert5, col sep=space] {summarization_zero_cost.csv};
            \addplot [darkorange1, thick, dashdotted, smooth] table [x=idx, y=mpnet1, col sep=space] {summarization_zero_cost.csv};
            \addplot [mpnet5, mark=x, thick, smooth] table [x=idx, y=mpnet5, col sep=space] {summarization_zero_cost.csv};
            \addplot [black, dotted, thick, smooth] table [x=idx, y=oracle, col sep=space] {summarization_zero_cost.csv};

            \addlegendimage{bert5}
            \addlegendentry{CIPHER-5-B}
            \addlegendimage{darkorange1}
            \addlegendentry{CIPHER-1-M}
            \addlegendimage{mpnet5}
            \addlegendentry{CIPHER-5-M}
            \addlegendimage{black}
            \addlegendentry{Oracle}

       \nextgroupplot[
              title=\textbf{Email Writing},title style={yshift=-1.5ex},
              ylabel={\% Zero-cost Ex. per Bin}, ymin=0.5, ymax=1,
              ylabel style={yshift=-2.5ex},
              xlabel={Round},xmin=0, xmax=9, xtick distance=1.8, xticklabels={-40,,40,80,120,160,200},
              xlabel style={yshift=1ex},
              yticklabels={,0,0.2,0.4,0.6,0.8,1},
              ]
            \addplot [bert5, mark=o, thick, smooth] table [x=idx, y=bert5, col sep=space] {email_zero_cost.csv};
            \addplot [darkorange1, thick, dashdotted, smooth] table [x=idx, y=mpnet1, col sep=space] {email_zero_cost.csv};
            \addplot [mpnet5, mark=x, thick, smooth] table [x=idx, y=mpnet5, col sep=space] {email_zero_cost.csv};
            \addplot [black, dotted, thick, smooth] table [x=idx, y=oracle, col sep=space] {email_zero_cost.csv};

\end{groupplot}
\end{tikzpicture}
\vspace{-20pt}
\label{fig:zero_cost}
\end{figure*}

 \noindent\textbf{Learning Curves.} We plot mean cumulative user edit costs over rounds in~\pref{fig:cumulative_cost}. The cumulative user edit costs in~\pref{fig:cumulative_cost} show that the angle of the learning curves decreases for \algname~after an initial number of rounds, showing that learning helps decrease the rate at which user edits are accumulated. In contrast, the angle of the learning curve for the no-learning baseline remains unchanged.\looseness=-1

 \noindent\textbf{Evaluating Fraction of Non-Edited Responses.} Recall that the first stage of our GPT-4 user checks if the agent response satisfies the latent user preference $f^\star$. If it does, then no edits are performed, otherwise, the user edits the response. We plot the percentage of examples with zero edit cost per 20 rounds bin in~\pref{fig:zero_cost}. We notice a small increase in the number of examples with zero edit cost. This indicates that gains come not just by increasing the number of examples that avoid getting edited in stage 1 of our user but more generally across examples.\looseness=-1

 \noindent\textbf{Qualitative Analysis of Learned Preferences.} We  evaluate the quality of preferences learned by~\algname~on the harder summarization task.
 \pref{tab:learned_prefs} lists 3 learned preferences per document source for \textit{\algname-5-MPNET} which are randomly sampled at the beginning, middle, and end of the interaction history. We see that overall the agent can learn a reasonable description of the latent preference. For example, it can learn \emph{bullet points} preference for Wikipedia articles, and \emph{second person narrative} for Reddit posts, and \emph{QA style} for Movie reviews. \algname~can pick some preferences fairly early such as \emph{bullet points} for Wikipedia and \emph{emojis} for Paper abstract, whereas some are learned only later such as \emph{Structured Q$\&$A} for Movie reviews. This shows using \algname~can quickly learn useful preferences, but further interaction continues to help.\footnote{We present more additional analysis in~\pref{app:additional_analysis}, including detailed expense report, normalized edit distance cost, failure case analysis, and retrieval accuracy.}\looseness=-1

\begin{table}[t!]
    \vspace{-15pt}
    \centering \small
    \setlength{\tabcolsep}{3pt}
    \caption{Examples of learned preferences on summarization task with \textit{CIPHER-5-MPNET}, grouped based on the document source and corresponding latent preference. We randomly sample 3 examples per type at the beginning, middle, and end of the interaction history.}   
    \begin{tabular}{p{0.25\linewidth} p{0.72\linewidth}} %
        \toprule
        \textbf{Latent User Preference} & \textbf{(Round) Learned Preference} \\
        \midrule
         \textbf{News article.} targeted to young children, storytelling, short sentences, playful language, interactive, positive & (22) Fairy tale narrative style, informal and conversational tone, use of rhetorical questions, simplified language.
         \newline (115) Simplified, childlike storytelling with playful language and imagery
         \newline (192) Simplified and playful storytelling language  \\
        \midrule
        \textbf{Reddit post.} second person narrative, brief, show emotions, invoke personal reflection, immersive   &  (14) Concise and coherent storytelling 
         \newline (102) The user prefers a second-person narrative and a more direct, personal tone 
         \newline (194) Poetic and descriptive language, narrative perspective shift to second person\\
         \midrule
         \textbf{Wikipedia page.} bullet points, parallel structure, brief & (19) Concise, Bullet-Pointed, Structured Summaries with a Narrative Q\&A Style
         \newline (124) Concise and factual writing style, bullet-point formatting 
            \newline (197) Concise and streamlined formatting, with bullet points and clear subheadings for easy scanning \\
         \midrule
         \textbf{Paper abstract.} tweet style, simple English, inquisitive, skillful foreshadowing, with emojis & (20) Concise, conversational summaries with bullet points and emojis. 
         \newline 
         (111) Concise, conversational, whimsical bullet-point summaries with emojis. \includegraphics[height=10pt]{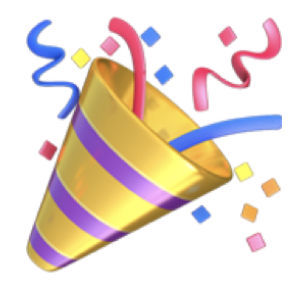} \includegraphics[height=10pt]{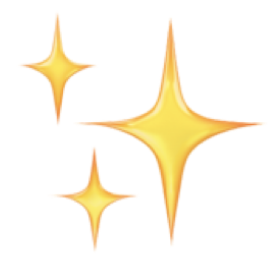} \includegraphics[height=10pt]{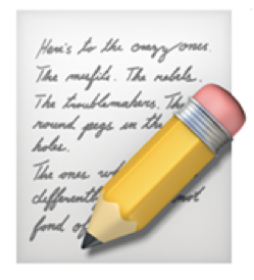}
                \newline
                    (193) Concise, conversational, and whimsical bullet-point summaries with emojis. \includegraphics[height=10pt]{graphs/tada.png} \includegraphics[height=10pt]{graphs/sparkles.png} \includegraphics[height=10pt]{graphs/write.png} \includegraphics[height=10pt]{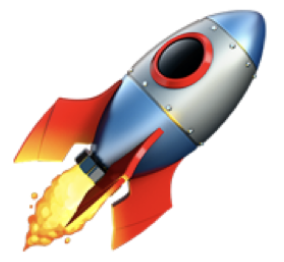} \\
        \midrule
        \textbf{Movie review.} question answering style & (12) The user prefers a straightforward, clear, and concise writing style with factual formatting.
        \newline (123) The user prefers a clear and concise question and answer format with straightforward language. 
        \newline (199) Concise, Structured Q\&A with Whimsical Clarity \\

        \bottomrule
    \end{tabular} 
    \label{tab:learned_prefs}
    \vspace{-15pt}
\end{table}

\subsection{Human Evaluation} 

We conduct two types of evaluation with human users to further understand the performance of our methods on summarization. We focus on our best-performing method \textit{CIPHER-5-MPNET}.\footnote{Examples in each evaluation have no overlap, which are sampled from experiments on different seeds.} 

\noindent\textbf{Win Rate Evaluation.} 
We conduct win rate evaluation where evaluators are given a pair of text and choose which one has higher quality. We compare the output of \textit{CIPHER-5-MPNET} against the output of the best-performing baseline~\textit{ICL-edit-5-MPNET}, and against the generation of the oracle method. Each evaluation covers 15 text pairs, with three random samples from each scenario in the last 50 rounds of interaction. We conduct these~\textit{CIPHER vs. ICL} and \textit{CIPHER vs. Oracle} evaluations with 7 human evaluators recruited through our personal network. For each text pair, we consider the output receiving the majority vote as a win.
We find that the win rate of \textit{CIPHER-5-MPNET} against \textit{ICL-edit-5-MPNET} is 73.3\%. This confirms that our method outperforms the best-performing baseline for human users.
In \textit{CIPHER vs. Oracle} evaluation, the win rate of \textit{CIPHER-5-MPNET} is 23.7\%, which reflects the performance gap we reported in previous sections. \looseness=-1

\noindent\textbf{Edits by Human Users.}
We study the edit feedback from human users to the generation of \textit{CIPHER-5-MPNET} and the oracle method. We instruct human users to edit the output based on the given latent preference, and to leave no edits when the output aligns with the given preference. We mix 20 outputs from \textit{CIPHER-5-MPNET} and the oracle method so that human users cannot tell the source of each output. The total edit distance, averaged across 3 human users, is 211 for CIPHER, and 98 for the oracle method. The averaged percentage of zero-edit examples is 60\% for CIPHER and 76.7\% for oracle. \looseness=-1

\section{Conclusion}
We study aligning LLM-based agents using user edits that arise naturally in applications such as writing assistants. We introduce the~\framework~framework that seeks to learn the latent user preferences that drive these edits, and uses them to generate a response.
We propose a practical algorithm~\algname~that implements~\framework and outperforms baselines on two interactive tasks with a GPT-4 simulated user. 
Evaluating~\algname~with human-in-the-loop as well as developing algorithms that can fine-tune LLMs using user edit where fine-tuning is feasible, are interesting future work directions.\looseness=-1

\subsubsection*{Acknowledgments}
Gao was a research intern in MSR NYC, and later was partially  supported by NSF project \#1901030. All content represents the opinion of the authors, which is not necessarily shared or endorsed by their respective employers and/or sponsors.
We thank MSR NYC research community, Jonathan D. Chang, Daniel D. Lee, Claire Cardie, and Sasha Rush for helpful discussions and support. We also thank Stéphane Aroca-Ouellette, Kyunghyun Cho, and Columbia NLP community for their valuable feedback.

\bibliography{neurips_2023}
\bibliographystyle{iclr2024_conference}

\newpage
\appendix
\section*{{\Large Appendix}}

\section{Related Work}

We describe related work in this area grouped by main themes in this work.

\paragraph{Learning from Feedback.} Besides pair-wise comparison feedback from annotators used in Reinforcement Learning from Human Feedback (RLHF) research~\citep[inter alia]{Ziegler2019FineTuningLM,Stiennon2020LearningTS,Nakano2021WebGPTBQ,Ouyang2022TrainingLM}, prior work has also studied free-form text feedback provided by annotators ~\citep{Fernandes2023BridgingTG}, such as on the task of dialog~\citep{Weston2016DialogbasedLL,Li2016DialogueLW,Hancock2019LearningFD,Xu2022LearningNS,Petrak2023LearningFF}, question answering~\citep{Li2022UsingIF,Malaviya2023PachinkoPI}, summarization~\citep{Saunders2022SelfcritiquingMF}, and general decision making~\citep{cheng2023llf}. This feedback, tailored to each example, is often utilized to rank candidate outputs, thereby improving task performance. Some work studies learning from text feedback to generate outputs directly~\citep{Scheurer2023TrainingLM,Bai2022ConstitutionalAH,Shi2022WhenLG}, by generating multiple refinements of the original output based on the feedback and fine-tuning the original model to maximize the likelihood of the best refinement. In grounded settings such as instruction-based navigation, one line of work has also used hindsight feedback that explicitly provides a text instruction for the generated trajectory, to train policies~\citep{nguyen2021interactive,misra2024provable}. Moving beyond the conventional focus on text feedback that explicitly articulates human intent, we investigate feedback in the form of direct edits on the original model output. Such revisions by users occur naturally during model deployment in practice.  Additionally, we examine the learning of user preferences through historical interactions, aiming to surpass the constraints of example-specific feedback.

\paragraph{Language Agents and Personalization.} LLMs have enabled the development of language agents for a variety of tasks from writing assistants~\citep{Lee2024ADS}, coding assistants~\citep{githubcopilot}, and customer service assistants~\citep{brynjolfsson2023generative}. Since these LLM-based assistants are often used by individuals, a natural question has arisen on how to personalize these agents for each user. Straightforward approaches for fine-tuning LLMs includes supervised learning, online DPO~\citep{guo2024direct}, learning-to-search~\citep{chang2023learning}, and reinforcement learning~\citep{ouyang2022training}. These approaches can be directly applied to our setting. For example, one can use $(y_t, y'_t)$ in~\pref{proto:learning_from_edits} as the preference data where $y'_t$ is preferred over $y_t$, or use $y'_t$ as the ground truth for supervised learning. However, fine-tuning is expensive and hard to scale with the number of users. Therefore, a line of work has explored improving the alignment of frozen LLMs by \emph{prompt engineering}, such as learning a personalized retrieval model~\citep{mysore2023pearl}, learning a prompt policy given a reward function~\citep{deng2022rlprompt}, or more generally, learning to rewrite the entire prompt~\citep{li2023automatic}. We focus on learning a prompt policy by learning from user edits, and specifically, using them to extract textural descriptions of user preference.\looseness=-1

\paragraph{Edits and Revisions.} Many prior work on editing model output focuses on error correction, such as fixing source code~\citep{Yin2018LearningTR,Chen2018SequenceRSL,Reid2023DiffusERDV} and improving the factual consistency of model summaries~\citep{Cao2020FactualEC,Liu2022OnIS,Balachandran2022CorrectingDF}. A line of work has explored understanding human edits based on edit history of Wikipedia~\citep{Botha2018LearningTS,Faltings2020TextEB,Rajagopal2022OneDM,Reid2022LearningTM,Laban2023SWiPEAD}, or revisions of academic writings~\citep{Mita2022TowardsAD,Du2022UnderstandingIR,DArcy2023ARIESAC}. Prior work explores predicting text revisions with edit intents~\citep{Brody2020ASM,Kim2022ImprovingIT,Chong2023LeveragingPT}, and modeling edits with various approaches, including latent vectors~\citep{Guu2017GeneratingSB,MarreseTaylor2020VariationalIF,MarreseTaylor2023EditAR}, structured trees~\citep{Yao2021LearningSE}, discrete diffusion process~\citep{Reid2023DiffusERDV}, or a series of singular edit operations~\citep{Stahlberg2020Seq2EditsST,Mallinson2020FELIXFT,Agrawal2022AnIL,Zhang2022CoditT5PF,Liu2023SecondTA}. However, these methodologies predominantly target generic improvements in model performance, overlooking the intricacies of individual user satisfaction and preference. Our research takes a distinct direction, focusing on understanding edits across a variety of examples to study user-level preferences, with a practical goal of aligning the agent to individual preferences.

\section{Additional Details}
\label{app:addition_details}

\paragraph{Dataset Examples.} We list links to dataset sources for our user-provided context in~\pref{tab:context_link_example}.

\paragraph{GPT-4 User's Edits} We list examples of OUR GPT-4 user's edits with different latent preference on summarization in~\pref{tab:user_edits}.

\paragraph{GPT-4 User Templates.} Prompt templates used by our GPT-4 user are provided in~\pref{tab:ai_user_prompt_template}.

\paragraph{Baseline Hyperparameters.} For E-then-e LPI and Continual LPI we set $T_e = 5$. 
For ICL-edit baselines, we experimented with different values of $k$, and report our best results with $k=5$.

\paragraph{\algname~Templates.} Prompt templates used by~\algname~are provided in~\pref{tab:agent_prompt_template}.

\paragraph{ICL-edit Templates.} Prompt templates used by \textit{ICL-edit} baseline are provided in~\pref{tab:icl-pref_prompt_template}.

\paragraph{CoT-edit Templates.} Prompt templates used by \textit{CoT-edit} baseline are provided in~\pref{tab:cot-pref_prompt_template}.

\section{Additional Analysis}
\label{app:additional_analysis}

\paragraph{Evaluating Normalized Edit Cost.} The cumulative user edit cost measures the total effort of the user but is susceptible to outlier examples, as the edit distance for a given round is potentially unbounded. Therefore, we also compute a \emph{normalized edit distance} $\cost(y_t, y'_t) / |y_t|$ by dividing the edit distance by $\max\{|y_t|, |y'_t|\}$, i.e. the max length of the agent output or user revised text. As Levenshtein distance $\cost(y_t, y'_t)$ is upper bounded by $\max\{|y_t|, |y'_t|\}$, therefore, the normalized cost is at most 1. \pref{fig:normalized_cost_appendix}~reports normalized cost over rounds for the top 3 methods. %
We notice that for all variants of~\algname~for the summarization task, and for~\algname-5-M for the email writing task, the normalized cost decreases notably as training progresses indicating learning. As the cost is normalized by the response length, even a small decrease can lead to a significant reduction in the number of tokens edited.

\paragraph{Detailed Expense.} We list a detailed computational expense of different methods in~\pref{tab:expense_breakdown}.

\paragraph{Failure Case Analysis.} ~\algname~notably reduces the edit cost and learns useful preference, however, significant gaps to the oracle method remain, especially in the summarization task. 
We manually analyze failure cases on summarization task with the best performing method \textit{CIPHER-5-MPNET}. ~\pref{tab:failures} in the Appendix reports the summary and example of our findings, categorized as preference inference from output-revision pair, consolidation of inferred preferences, and retrieval. In brief, the most common type of failure is on the preference inference step given the agent output and user revision. For example, the agent often misses the exact keyword for \textit{brief} or \textit{short sentences}, and sometimes struggles with inferring the \textit{second-person narrative} aspect.

\paragraph{Retrieval Accuracy.} We calculate retrieval accuracy for ~\algname~as the fraction of all retrieved
contexts that are of the same document type as the currently given context across all seeds and time steps.  We report the results in~\pref{tab:retrieval_acc}. We find that the retrieval accuracy is higher on the summarization task than on email writing. and using MPNET typically performs better than using Bert to encode context.

\paragraph{Survey Details.} We did a small survey with several participants recruited from our personal network. The instructions for the two tasks are as follows:
\begin{enumerate}
    \item Task 1 instruction: \emph{``You're asked to compare 2 pieces of writing in terms of satisfaction towards certain preference. There are 15 pairs to compare in total."}. For a specific example, we ask \emph{``Assume that the writing style you prefer is <preference>. (e.g., you want to quickly get main opinions from a movie review) Which piece of writing below do you like better"}. We replace <preference> with the given preference.
    \item Task 2 instruction: \emph``{This study is a simulation of how you use AI writing assistants. There are 20 pieces of text that you need to review. You will be given an assumption of the preferred writing style, and you can edit the piece of writing if it doesn’t satisfy the given preference. If the writing aligns with the specified style, please leave no edits! Please treat each writing as an independent piece, even though some writings are based on the same article. For example, when reviewing the writing and making edits, do not bring in the knowledge you learned from another piece of writing.}
\end{enumerate}

\begin{table}[t!]
    \centering \small
    \caption{Link to each source dataset, from which we randomly sample examples as the user-provided context in our tasks.} 
    \setlength{\tabcolsep}{0.01\linewidth}
    \begin{tabular}{p{0.3\linewidth} p{0.67\linewidth}}
        \toprule
        \textbf{Data Source} & \textbf{Link and Example} \\
        \midrule
        CNN Daily Mail \citep{see-etal-2017-get} & \url{https://huggingface.co/datasets/cnn_dailymail} \\
        SLF5K  \citep{Stiennon2020LearningTS} & \url{https://huggingface.co/datasets/JeremyAlain/SLF5K} \\
        Wikidump  \citep{wikidump} & \url{https://huggingface.co/datasets/wikipedia} \\
        Arxiv \citep{clement2019arxiv} & \url{https://huggingface.co/datasets/CShorten/ML-ArXiv-Papers} \\
 
        IMDb \citep{maas-EtAl:2011:ACL-HLT2011} & \url{https://huggingface.co/datasets/imdb}  \\
        PaperTweet \citep{Bar_PaperTweet} & \url{https://huggingface.co/datasets/nitsanb/paper_tweet} \\
         Elsevier OA CC-By \citep{Kershaw2020ElsevierOC} & \url{https://huggingface.co/datasets/orieg/elsevier-oa-cc-by} \\
 
        \bottomrule
    \end{tabular}   
    \label{tab:context_link_example}
\end{table}

\begin{table}[h!]
    \centering \small
    \caption{Examples of user edits with different latent preference on summarization. The agent output column shows the agent generation without conditioning on any preference. The latent preference for the user edits is in bold and parenthesis. We truncate some examples due to limited space. }
    \begin{tabular}{p{0.45\textwidth} p{0.45\textwidth}}
        \toprule
        \textbf{Agent Output}  & \textbf{User Edits} \\
        \midrule
Leon Fenner, the suspected gunman in a \$2 million diamond heist at Watch Standard Jewelry in New York City, was arrested in Maryland. The robbery occurred on November 11 and resulted in the injury of the store owner's elderly father. Fenner's accomplice, Rondu Frisby, was a friend of the store owner and was apprehended a week after the heist. The robbery took place during the city's Veteran's Day Parade, leading to a block shutdown. Fenner and Frisby stole numerous jewels and 20 luxury watches, but were caught on surveillance cameras. Fenner is set to appear before a federal court in Maryland. 
& \textbf{(targeted to young children, storytelling, short sentences, playful language, interactive, positive)} \newline
Once upon a time in the big, bustling city of New York, during a parade filled with music and marching on Veteran's Day, something sneaky happened at a place called Watch Standard Jewelry. This wasn't just any store; it was a treasure chest full of sparkling diamonds and shiny watches that could make your eyes twinkle like stars.

Now, two not-so-nice men, Leon Fenner and his buddy Rondu Frisby, thought they could be sneaky and take these treasures without asking. On November 11, while everyone was watching the parade, they tiptoed into the store. But oh no! They accidentally hurt the store owner's dear old dad, which wasn't very kind at all.
\textit{[TRUNCATED]} \\

\midrule
The article describes a situation where a neighbor's cat, one of seven rescued by a "cat lady," has started frequenting the author's front yard after leaving its original home due to the presence of other cats. Over the past few months, the cat has become more consistent in its visits, meowing at the author and their family members for attention. The cat allows itself to be petted and shows affectionate behavior, such as rubbing against legs and headbutting hands, until it is fed. After eating, the cat typically leaves but returns in the evening. The author is seeking advice on how to handle the situation, expressing concern about feeding the cat because it is not legally theirs and there could be potential ownership and legal issues.
& \textbf{(second person narrative, brief, show emotions, invoke personal reflection, immersive)} \newline
Imagine yourself in a quaint neighborhood, where the gentle purrs of a feline friend have become a familiar part of your daily routine. This isn't just any cat—it's one of seven, a charming escapee from the home of a local ""cat lady,"" seeking solace in the tranquility of your front yard. Over recent months, this whiskered wanderer has turned your doorstep into its personal stage, meowing for attention with a persistence that tugs at your heartstrings.

Each encounter with this affectionate creature is a dance of mutual admiration. You find yourself captivated by its soft fur and the way it eagerly rubs against your legs, headbutting your hands in a silent plea for affection.
\textit{[TRUNCATED]} \\

\midrule

The Puerto Plata Province is a northern province in the Dominican Republic, known for its beautiful beaches and as a popular tourist destination since the late 1990s. Its capital is San Felipe de Puerto Plata, named after the silver clouds Columbus saw on a mountain nearby in 1493. The city of Puerto Plata was founded in 1502.

Historically, the first European town in the Americas, La Isabela, was established by Columbus near Puerto Plata in 1493. In 1605, the governor Antonio de Osorio destroyed Puerto Plata and other towns to curb illegal trade, leading to the formation of new settlements.

Puerto Plata became a Maritime District in 1865 and was later recognized as a province in the 1907 constitution. Over time, it expanded to include more municipalities and municipal districts, with the latest additions in 2007.
\textit{[TRUNCATED]}

    & \textbf{(bullet points, parallel structure, brief)} \newline
- Location: Northern Dominican Republic \newline
- Capital: San Felipe de Puerto Plata \newline
- Founding: 1502 by Europeans\newline
- Historical site: La Isabela (1493, first European town in the Americas)\newline
- 1605: Destruction by Governor Antonio de Osorio to stop illegal trade\newline
- Status: Became Maritime District in 1865, recognized as province in 1907\newline
- Expansion: Added more municipalities, latest in 2007\newline
- Borders: Atlantic Ocean, Espaillat, Santiago, Valverde, Monte Cristi provinces\newline
\textit{[TRUNCATED]} \\
\midrule
\end{tabular}
    \label{tab:user_edits}
\end{table}
\begin{table}[h!]
    \centering \small
    \begin{tabular}{p{0.45\textwidth} p{0.45\textwidth}}
        \midrule

The article discusses the challenge of detecting and classifying stuttered speech, which affects many people but has been subject to limited research. The authors present a new model that differs from most existing methods, which typically use language models for stutter identification. Instead, their model is based on acoustic features, which allows it to identify various kinds of stuttering disfluencies without relying on speech recognition technology. The proposed model combines a deep residual network with bidirectional long short-term memory (LSTM) layers to classify different stutter types. It has been shown to significantly outperform current leading methods, reducing the average miss rate to 10.03\%, which is a 27\% improvement over the state-of-the-art.
& \textbf{(tweet style, simple English, inquisitive, skillful foreshadowing, with emojis)} \newline 
Ever struggled with stuttering or know someone who does? \includegraphics[height=10pt]{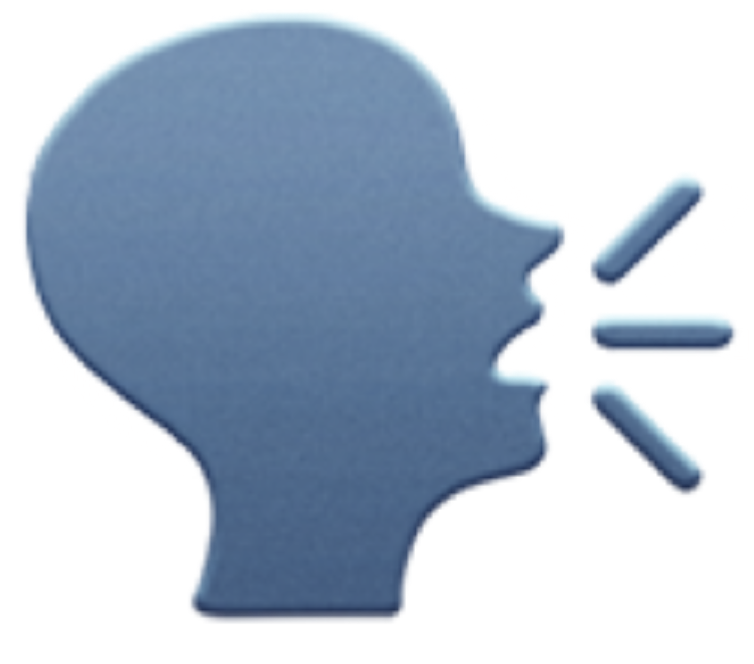} Research on detecting stutters hasn't been huge, but guess what? A new study is changing the game! \includegraphics[height=10pt]{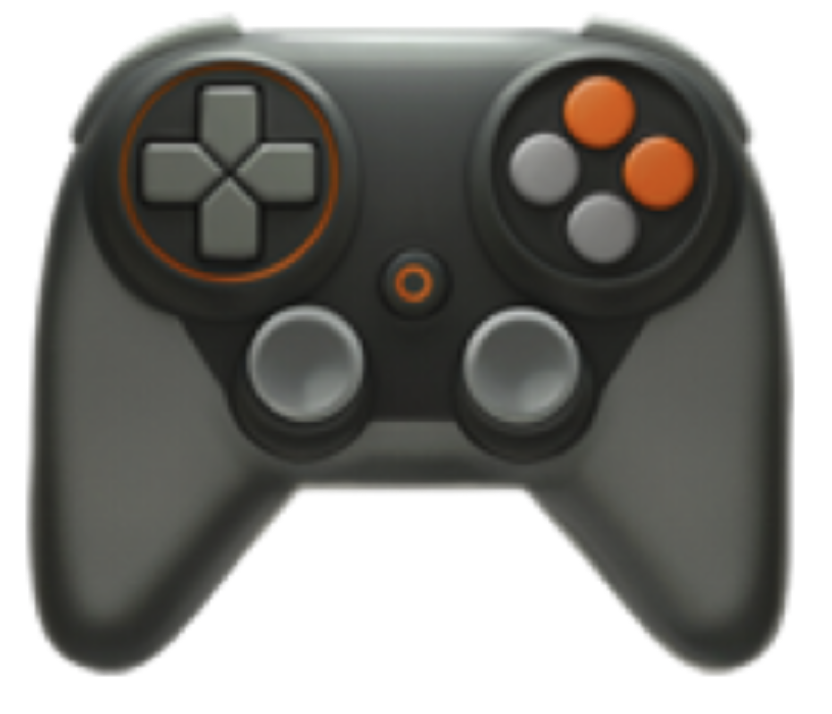} Instead of just looking at words, this smart model listens to how things are said. \includegraphics[height=10pt]{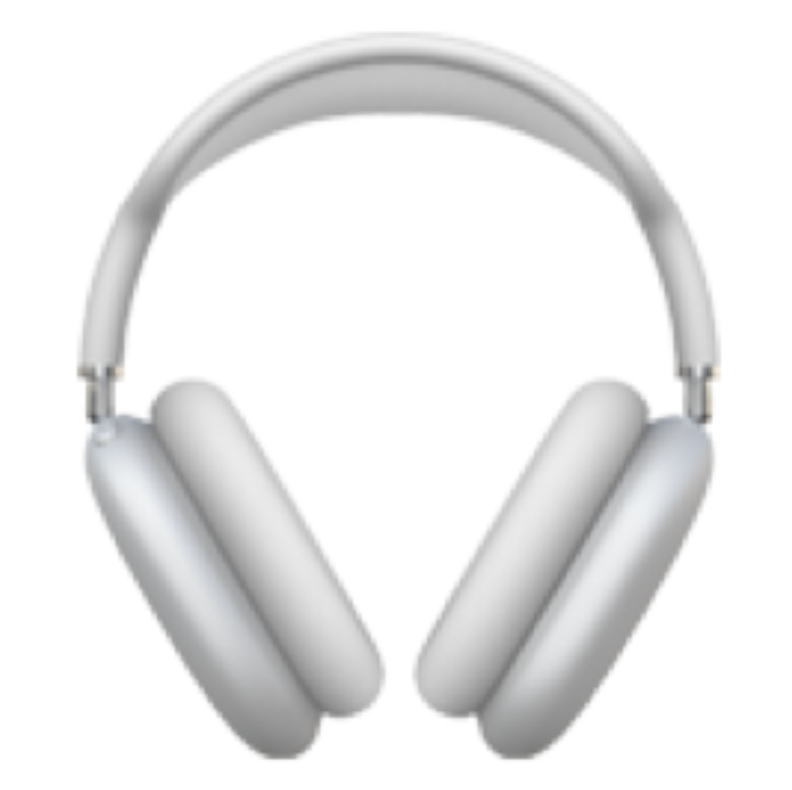} It's like a detective for sounds, using cool tech called deep learning and memory networks to spot different stutter types. \includegraphics[height=10pt]{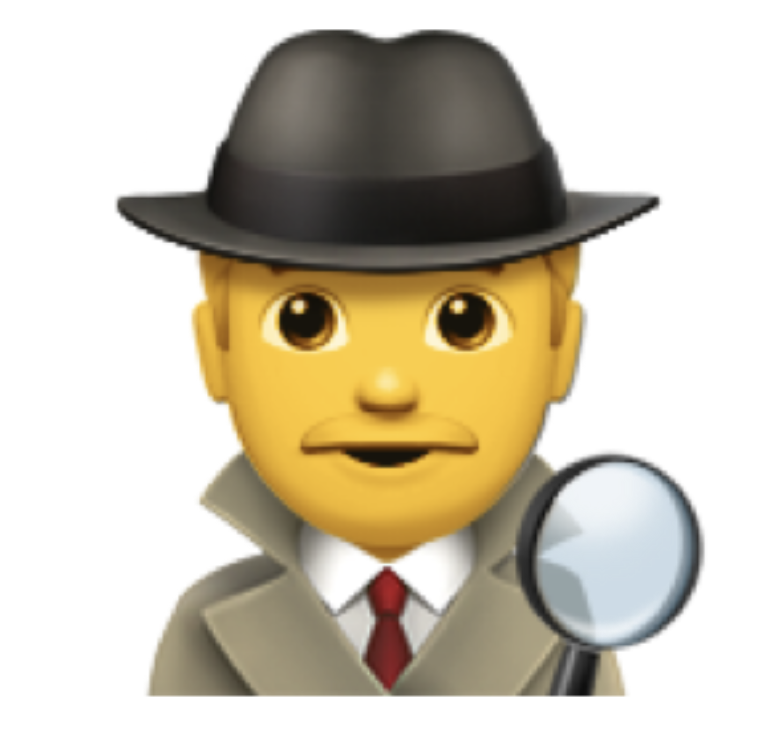} And it's crushing it, beating old methods by 27\%! Missed stutters? Down to just 10\%! \includegraphics[height=10pt]{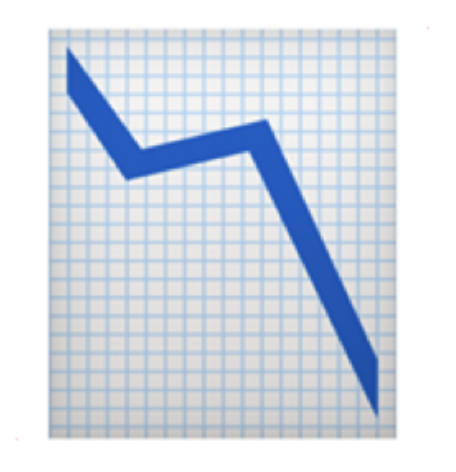} This could be a big deal for speech therapy! \includegraphics[height=10pt]{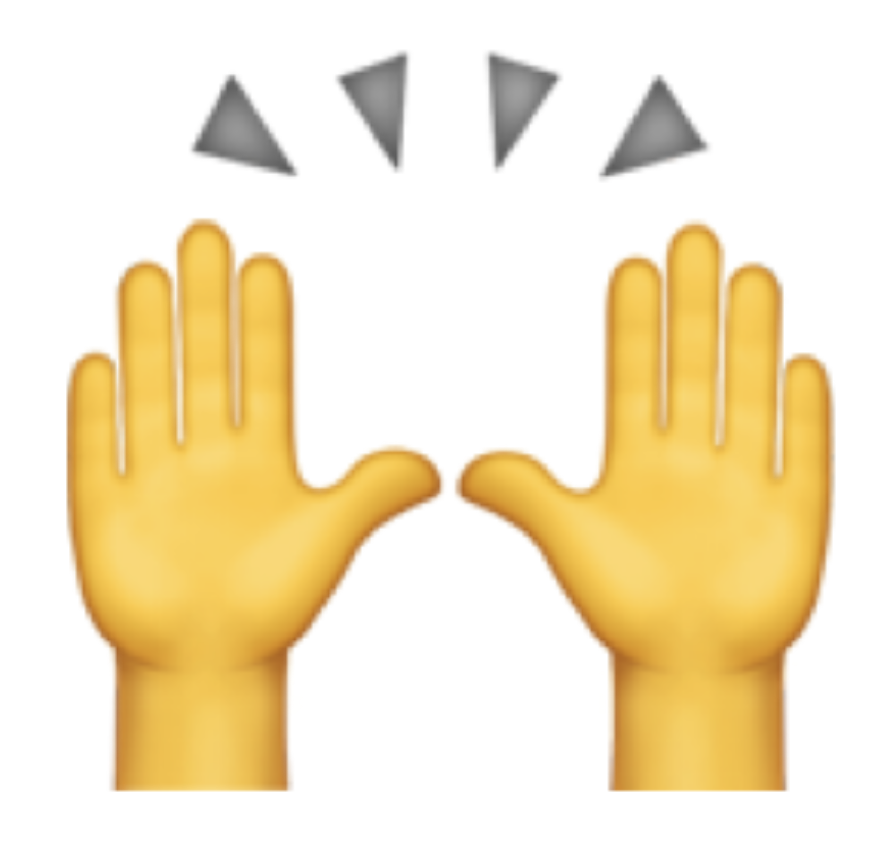} \#Stuttering \# SpeechTech \# Innovation \\
\midrule
The article discusses a movie that begins promisingly but later becomes confusing due to a poorly depicted portrayal of ancient black magic rituals. The male lead delivers a strong performance despite his character's unfavorable outcome in the climax. The movie is compared to a mix of "Boomerang" and "Extremities."
    & \textbf{(question answering style, direct, concise)} \newline
What is the general sentiment of the article about the movie?

The article suggests that the movie starts off well but eventually becomes muddled. The depiction of ancient black magic rituals is criticized for being confusing.

How does the article describe the male lead's performance?

The male lead's performance is described as strong, even though his character does not have a favorable outcome in the climax.

What films is the movie compared to in the article?

The movie is compared to ""Boomerang"" and ""Extremities"" in the article. 
\\

        \bottomrule
    \end{tabular}
\end{table}

\begin{table}[h!]
\centering \small
\caption{Prompt templates for the AI user. The first step is to prompt the user for yes/no answer regarding satisfaction. If the answer is no, the second step is to ask the user edit the agent output according to the latent preference. If the answer is yes, the agent output receives 0 edits.}
\begin{tabular}{p{0.07\linewidth} p{0.4\linewidth} p{0.44\linewidth}}
\toprule
     & \textbf{Summarization} & \textbf{Email Writing} \\
\midrule
    Step 1
    & Article: \context{\{user-provided article\}} \newline
        Summary: \response{\{agent-generated summary\}} \newline
        Is the above summary of the above article good for person who would love to use the following style: \preference{\{latent user preference\}}? Please answer yes or no. 
    & Notes: \context{\{user-provided notes\}} \newline
        Email: \response{\{agent-generated email\}} \newline
        Is the above email based on the above notes good for a user who wants the following style: \preference{\{latent user preference\}}? Please answer yes or no. \\
\midrule
    Step 2
    & Summary: \response{\{agent-generated summary\}}  \newline
        Please revise the above summary of an article to meet your style: \preference{\{latent user preference\}}. 
    & Email: \response{\{agent-generated email\}}  \newline
        Assume that you prefer \preference{\{latent user preference\}}. 
        Please revise the above email to meet your style. \\
\bottomrule
\end{tabular}
    \label{tab:ai_user_prompt_template}
\end{table}

\begin{table}[h!]
\centering \small
\vspace{-10pt}
\caption{Prompt templates for CIPHER.}%
\begin{tabular}{p{0.12\linewidth} p{0.42\linewidth} p{0.42\linewidth}}
\toprule
     & \textbf{Summarization} & \textbf{Email Writing} \\
\midrule
    Task prompt conditioned on inferred preference \newline (\pref{line:generate} in~\pref{alg:cipher})
    & Article: \context{\{user-provided article\}} \newline
        Assume that you need to summarize the above article for a user, 
        who prefers the following style: \preference{\{inferred user preference\}}. 
        Please write a summary of the above article to address those specified preferences.
    & Notes: \context{\{user-provided notes\}} \newline
        These notes are written by a user who prefers the following style of emails: \preference{\{inferred user preference\}}. 
        Please write a short email based on the above notes to address those specified preferences. \\
\midrule
Prompt to infer user preference based on revision \newline (\pref{line:infer_preference} in~\pref{alg:cipher})
    & Original summary of an article: \response{\{agent-generated summary\}} \newline
        Revised summary by a user: \revision{\{user revision\}} \newline
        Based on the edits and revision by this user on the original summary in the above examples, 
        what do you find about this user's generic preference in terms of writing style and formatting?  
        Please answer in a short phrase and only recommend those preferences that are widely used.
    & Original email: \response{\{agent-generated email\}} \newline
        Revised email: \revision{\{user revision\}} \newline
        Based on the edits and revision by this user on the original email in the above examples, 
        what do you find about this user's generic preference in terms of writing style and formatting?  
        Please answer in a short phrase and only recommend those preferences that are widely used. \\
\midrule
    Prompt to consolidate inferred preferences from history \newline (\pref{line:merge_preferences} in~\pref{alg:cipher})
    & List of user preferences successfully being used to generate summaries of similar documents: \newline
        - \preference{\{inferred preference in a retrieved example\}} \newline
        - \preference{\{inferred preference in a retrieved example\}} \newline
        ... \newline
        Based on the the above examples, please come up with short phrase with the most represented summarization preferences of the user. 
    & List of user preferences successfully being used to generate emails of a similar kind: \newline
        - \preference{\{inferred preference in a retrieved example\}} \newline
        - \preference{\{inferred preference in a retrieved example\}} \newline
        ... \newline
        Based on the the above examples, please come up with short phrase with the most represented writing preferences of this user. \\
        
\bottomrule
\end{tabular}
    \label{tab:agent_prompt_template}
\end{table}

\begin{table}[h!]
\vspace{-10pt}
\centering \small
\caption{Prompt templates for the \textit{ICL-edit} baseline.}
\begin{tabular}{p{0.12\linewidth} p{0.42\linewidth} p{0.42\linewidth}}
\toprule
     & \textbf{Summarization} & \textbf{Email Writing} \\
    \midrule
    Prompt with retrieved user edit examples
    & Original summary of an article: \response{\{agent-generated summary in a retrieved example\}} \newline
        Revised summary by a user: \revision{\{user revision in a retrieved example\}} \newline
        Original summary of an article: \response{\{agent-generated summary in a retrieved example\}} \newline
        Revised summary by a user: \revision{\{user revision in a retrieved example\}} \newline
        ... \newline
        Article: \context{\{user-provided article\}} \newline
        Based on the edits and revision by this user on the original summary in the above examples, 
        Please summarize the above article:
    & Original summary of an article: \response{\{agent-generated summary in a retrieved example\}} \newline
        Revised summary by a user: \revision{\{user revision in a retrieved example\}} \newline
        Original summary of an article: \response{\{agent-generated summary in a retrieved example\}} \newline
        Revised summary by a user: \revision{\{user revision in a retrieved example\}} \newline
        ... \newline
        Notes: \context{\{user-provided notes\}} \newline
        Based on the edits and revision by this user on the original email in the above examples, 
        Please write an email based on the above notes for this user: \\
\bottomrule
\end{tabular}
    \label{tab:icl-pref_prompt_template}
\end{table}

\begin{table}[h!]
\vspace{-10pt}
\centering \small
\caption{Prompt templates for the \textit{CoT-edit} baseline.}
\begin{tabular}{p{0.12\linewidth} p{0.42\linewidth} p{0.42\linewidth}}
\toprule
     & \textbf{Summarization} & \textbf{Email Writing} \\
    \midrule
    Prompt with retrieved user edit examples 
    & Original summary of an article: \response{\{agent-generated summary in a retrieved example\}} \newline
        Revised summary by a user: \revision{\{user revision in a retrieved example\}} \newline
        Original summary of an article: \response{\{agent-generated summary in a retrieved example\}} \newline
        Revised summary by a user: \revision{\{user revision in a retrieved example\}} \newline
        ... \newline
        FIRST, come up with short phrases that explain edits made by the user in the above examples, to show this user's writing preference. 
        SECOND, on a new line, summarize the above given article using the inferred preference.
        Your response MUST FOLLOW THE FOLLOWING FORMAT:
        PREFERENCE: <your inferred preference as a list of short phrases>
        RESULT: <your summary of the given article>
    & Original summary of an article: \response{\{agent-generated summary in a retrieved example\}} \newline
        Revised summary by a user: \revision{\{user revision in a retrieved example\}} \newline
        Original summary of an article: \response{\{agent-generated summary in a retrieved example\}} \newline
        Revised summary by a user: \revision{\{user revision in a retrieved example\}} \newline
        ... \newline
        Notes: \context{\{user-provided notes\}} \newline
        FIRST, come up with short phrases that explain edits made by the user in the above examples, to show this user's writing preference. 
        SECOND, on a new line, write an email based on the above notes using the inferred preference.
        Your response MUST FOLLOW THE FOLLOWING FORMAT:
        PREFERENCE: <your inferred preference as a list of short phrases>
        RESULT: <your email based on the given notes> \\
\bottomrule
\end{tabular}
    \label{tab:cot-pref_prompt_template}
\end{table}

\begin{table}[t!]
\centering \small

\setlength{\tabcolsep}{14pt}
\caption{Expense of different methods: number of BPE tokens in terms of input, output and total. Each number is the average across 3 runs (unit is $\cdot 10^5$). }
\begin{tabular}{l c c c c c c }
\toprule
    \textbf{Method} & \multicolumn{3}{c}{\textbf{Summarization}} & \multicolumn{3}{c}{\textbf{Email Writing}} \\
    & Input & Output & Total & Input & Output & Total \\
\midrule
    Oracle Preference & 1.14 & 0.53 & 1.67 & 0.91 & 0.71 & 1.62 \\
\midrule
    No Learning &  1.06 & 0.44 & 1.50 & 0.85 & 0.80 & 1.65 \\
    E-then-e LPI & 1.16 & 0.83 & 1.99 & 0.94 & 0.79 & 1.73 \\
    Continual LPI & 8.14 & 0.75 & 8.89 & 7.89 & 0.73 & 8.63 \\
\midrule
    ICL-edit-5-MPNET & 7.35 & 0.65 & 8.00 & 11.05 & 1.06 & 12.12 \\
    ICL-edit-5-BERT & 7.32 & 0.64 & 7.96 & 10.51 & 1.03 & 11.55 \\
    CoT-edit-5-MPNET & 6.23 & 0.59 & 6.82 & 7.67 & 0.79 & 8.47 \\
    CoT-edit-5-BERT & 6.34 & 0.58 & 6.92 &	7.48 & 0.78 & 8.26 \\
\midrule
    CIPHER-1-MPNET & 2.02 & 0.72 & 2.74 & 1.21 & 0.73 & 1.94 \\
    CIPHER-5-MPNET & 2.27 & 0.73 & 3.00 & 1.44 & 0.64 & 2.09 \\
    CIPHER-1-BERT & 2.10 & 0.71 & 2.81 & 1.27 & 0.73 & 1.99 \\
    CIPHER-5-BERT & 2.32 & 0.71 & 3.03 & 1.48 & 0.73 & 2.22 \\
\bottomrule
\end{tabular}
    \label{tab:expense_breakdown}
\end{table}

\begin{table}[h!]
    \centering \small
    \caption{Summary of failure cases on summarization task with \textit{CIPHER-5-MPNET}.}   
    \begin{tabular}{p{0.2\textwidth} p{0.25\textwidth} p{0.45\textwidth}}
        \toprule
        \textbf{Type of Failures} & \textbf{Summary} & \textbf{Examples}\\
        \midrule
        Preference inference based on an output-revision pair ($f_t$) \newline \textit{(the most common failure type)}
            & 1) Not totally wrong but insufficient, i.e. the inferred preference only captures a few aspects of user's latent preference. This is most common for news articles and Reddit posts, for which the user shows nuanced preference for several aspects.
            & The dominant missing aspect is \textit{brief} or \textit{short sentences} across different context, although the agent can infer keywords such as \textit{concise}. For news article context, the agent tends to infer the preference keyword \textit{whimsical}. The agent has difficulty to infer subtle aspects, including \textit{invoke personal reflection}, \textit{immersive}, \textit{positive}, \textit{parallel structure},  \textit{inquisitive}, and \textit{skillful foreshadowing}.\vspace{5pt}
            \\ 
            & 2) Sometimes fail to infer some important aspects, even though the user edits clearly show such preference.
            & The agent often could not infer \textit{second-person narrative}. For \textit{question answering style}, the agent occasionally only learns \textit{consistent format}.
            
            \\
        \midrule
        Consolidation of induced preferences from retrieved interactions ($\tilde{f}_t$) 
            & Overall, this step can capture the majority preference relatively well, although it tends to result in a more general preference compared to the retrieved ones.
            & When both specific phrase \textit{second-person narrative} and general phrase \textit{narrative} or \textit{narration} occur in retrieved examples, the agent often chooses to give a final preference not including the second-person perspective aspect. \\
        \midrule
        Retrieval of historical examples relevant to the given context 
            & The retrieval part in general works reasonably well, with more than half of the retrieved example being truly relevant to the given context. Note that one incorrect retrieved example typically does not affect the performance, as we instruct the agent to only use the most represented preference keywords among all five retrieved examples.
            & The agent sometimes retrieves wrong examples for Wikipedia context when its topic very relates to other context, e.g. wrongly retrieving past examples on news articles and movie reviews when the topic in the given Wikipedia context relates to these domains.
        \\
        \bottomrule

    \end{tabular}
\label{tab:failures}
\end{table}

\begin{table}[!h]
    \centering
     \caption{We report retrieval accuracy as the percentage of total retrieved document representations across all time steps and seeds that are of the same document source type as the context document for which they were retrieved. We use 3 seeds. We retrieve 600 examples for $k=1$ and 2970 examples for $k=5$.}
    \label{tab:retrieval_acc}
    \setlength{\tabcolsep}{0.01\linewidth}
    \begin{tabular}{lcc}
        \toprule
        \textbf{Method} & \textbf{Summarization} & \textbf{Email Writing} \\
        \midrule
        \algname-1-B & 72.00 &  25.83 \\ 
        \algname-1-M & \textbf{82.00} & \textbf{26.33} \\ 
        \midrule
        \algname-5-B & 65.79 & \textbf{26.57} \\ 
        \algname-5-M & \textbf{76.33} & 25.45\\         
        \bottomrule
    \end{tabular}
\end{table}

\begin{figure*}[t!]
\centering \small
\caption{Normalized cost of CIPHER over time, binned per 20 rounds to show the trend (average across 3 seeds). In the legend, \textit{-k} means with top $k$ retrieved examples, \textit{-B} for BERT, and \textit{-M} for MPNET.}
\vspace{-5pt}
\begin{tikzpicture}
\begin{groupplot}[
       group style={
          group size=2 by 2,
          vertical sep=25pt,
          horizontal sep=110pt,
          x descriptions at=edge bottom},
       ylabel style={align=center, text width=.22\textwidth},
       width=.4\textwidth,
       height=.27\textwidth],
       \nextgroupplot[
              title=\textbf{Summarization},title style={yshift=-1.5ex},
              ylabel={Normalized Cost}, ymin=0, ymax=0.7,
              ylabel style={yshift=-3.5ex},
              xmin=0, xmax=9, xtick distance=1.8,
              xlabel={Round},xticklabels={-40,,40,80,120,160,200},
              xlabel style={yshift=1ex},
              legend style={at={(1.05, 0.65)}, anchor=west, font=\scriptsize,, legend columns=1},
              ]
            \addplot [bert5, mark=o, thick, smooth] table [x=idx, y=bert5, col sep=space] {summarization_norm_cost.csv};
            \addplot [darkorange1, thick, dashdotted, smooth] table [x=idx, y=mpnet1, col sep=space] {summarization_norm_cost.csv};
            \addplot [mpnet5, mark=x, thick, smooth] table [x=idx, y=mpnet5, col sep=space] {summarization_norm_cost.csv};
            \addplot [black, dotted, thick, smooth] table [x=idx, y=oracle, col sep=space] {summarization_norm_cost.csv};

            \addlegendimage{bert5}
            \addlegendentry{CIPHER-5-B}
            \addlegendimage{darkorange1}
            \addlegendentry{CIPHER-1-M}
            \addlegendimage{mpnet5}
            \addlegendentry{CIPHER-5-M}
            \addlegendimage{black}
            \addlegendentry{Oracle}

       \nextgroupplot[
              title=\textbf{Email Writing},title style={yshift=-1.5ex},
              ylabel={Normalized Cost}, ymin=0, ymax=0.5,
              ylabel style={yshift=-3.5ex},
              xmin=0, xmax=9, xtick distance=1.8,
              xlabel={Round},xticklabels={-40,,40,80,120,160,200},
              xlabel style={yshift=1ex},
              legend style={font=\scriptsize}
              ]
            \addplot [bert5, mark=o, thick, smooth] table [x=idx, y=bert5, col sep=space] {email_norm_cost.csv};
            \addplot [darkorange1, thick, dashdotted, smooth] table [x=idx, y=mpnet1, col sep=space] {email_norm_cost.csv};
            \addplot [mpnet5, mark=x, thick, smooth] table [x=idx, y=mpnet5, col sep=space] {email_norm_cost.csv};
            \addplot [black, dotted, thick, smooth] table [x=idx, y=oracle, col sep=space] {email_norm_cost.csv};

\end{groupplot}
\end{tikzpicture}
\label{fig:normalized_cost_appendix}
\end{figure*}

\section{Broader Impact Statement}
Our work provides a way to develop language agents that learn from user edits. Our work is part of a broader effort to release language-based agents and consequently shares all risks associated with deploying language agents such as hallucination or bias innate in these models. We advocate for caution and thorough testing in releasing language agents. Further, we strongly emphasize seeking the permission of users before releasing language agents that learn from their edits. Specially, users should have the option to take back their consent at any time.

\end{document}